\documentclass[10pt,twocolumn,letterpaper]{article}

\usepackage[pagenumbers]{cvpr} 

\usepackage{graphicx}
\usepackage{amsmath}
\usepackage{amssymb}
\usepackage{booktabs}
\usepackage{comment}

\usepackage{pifont}
\newcommand{\xmark}{\ding{55}}%
\usepackage{multirow}
\usepackage{booktabs}
\usepackage{url}     
\usepackage{algorithm,algorithmic}
\usepackage{array}
\newcolumntype{P}[1]{>{\centering\arraybackslash}p{#1}}
\usepackage{stmaryrd}
\usepackage{subfiles}

\usepackage[pagebackref,breaklinks,colorlinks]{hyperref}

\usepackage[capitalize]{cleveref}
\crefname{section}{Sec.}{Secs.}
\Crefname{section}{Section}{Sections}
\Crefname{table}{Table}{Tables}
\crefname{table}{Tab.}{Tabs.}

\def\cvprPaperID{65} 
\def\confName{CVPR}
\def\confYear{2022}

\begin{document}

\title{Surface Analysis with Vision Transformers}

\author{Simon Dahan\\
King's College London\\
{\tt\small simon.dahan@kcl.ac.uk}
\and
Logan Z. J. Williams\\
King's College London\\
{\tt\small logan.williams@kcl.ac.uk}
\and
Abdulah Fawaz\\
King's College London\\
{\tt\small abdulah.fawaz@kcl.ac.uk}
\and
Daniel Rueckert\\
Imperial College London\\
{\tt\small d.rueckert@imperial.ac.uk}
\and
Emma C. Robinson\\
King's College London\\
{\tt\small emma.robinson@kcl.ac.uk}
}
\maketitle

\begin{abstract}

The extension of convolutional neural networks (CNNs) to non-Euclidean geometries has led to multiple frameworks for studying manifolds. Many of those methods have shown design limitations resulting in poor modelling of long-range associations, as the generalisation of convolutions to irregular surfaces is non-trivial. Recent state-of-the-art performance of Vision Transformers (ViTs) demonstrates that a general-purpose architecture, which implements self-attention, could replace the local feature learning operations of CNNs. Motivated by the success of attention-modelling in computer vision, we extend ViTs to surfaces by reformulating the task of surface learning as a sequence-to-sequence problem and propose a patching mechanism for surface meshes. We validate the performance of the proposed Surface Vision Transformer (SiT) on two brain age prediction tasks in the developing Human Connectome Project (dHCP) dataset
and investigate the impact of pre-training on model performance. Experiments show that the SiT outperforms many surface CNNs, while indicating some evidence of general transformation invariance. Code available at: \url{https://github.com/metrics-lab/surface-vision-transformers}

\end{abstract}
\section{Introduction}
\label{sec:intro}



Studying surfaces and manifolds is critical for many applications in computer graphics \cite{F.Monti2017,C.Qi2017,Litany2017}, protein interaction \cite{Gainza2019,Morehead2021} and biomedical analysis including cardiac \cite{C.Mauger2019, H.Xu2021}, brain \cite{B.Fischl2008,M.Glasser2013,M.Glasser2016,E.Robinson2014,E.Robinson2018,R.Dimitrova2021} imaging, with applications in biophysical and shape modelling \cite{cootes2001statistical,garcia2018dynamic,williams2010anatomically}, and mapping of cortical organisation \cite{M.Glasser2016,R.Dimitrova2021,L.Williams2021}.  While the shapes of meshes may vary greatly, ultimately, all problems may be reduced to analysis of functions over tessellated, deformable meshes. Despite this, there is no unified geometric deep learning method (gDL) for studying all these problems, and many gDL frameworks would in principle be suitable for estimating surface convolutions 
\cite{M.Bronstein2021, C.Qi2017, T.Cohen2018,M.Defferrard2017}. A recent work \cite{A.Fawaz2021} benchmarked various surface CNN methods on cortical phenotype regression and segmentation, and showed that they typically involve trade-offs between computational complexity, feature expressivity, and rotational equivariance of the models. 

While the locality and weight sharing properties of the convolution operation have pushed forward the computer vision field, especially in natural imaging, and created sample-efficient architectures that can generalise to a broad range of tasks \cite{K.He2015,T.Lin2016,M.Tan2019}, these inductive biases towards locality and scale-invariance in CNN architectures also induce a limited receptive field that impairs the modelling of long-range spatial dependencies between distant parts of an image \cite{M.Zeiler2013,K.Simonyan2015}. This prevents CNNs from efficiently modelling processes that are diffuse in space and/or time; something that is known to be true for a wide range of applications, specifically for biomedical applications where conditions may span over large areas \cite{BH.Kim2021,D.Perperidis2005,R.Dimitrova2021}.

Recently, Dosovitskiy et al. proposed the Vision Transformer (ViT), which sought to extend the use of self-attention transformer architectures, used in Natural Language Processing (NLP), to imaging data, by treating computer vision tasks as a sequence-to-sequence learning problem. By doing so, they showed that a general-purpose transformer architecture \cite{A.Vaswani2017} could be used for natural image classification; thereby demonstrating the benefits of using self-attention (SA) on image patches to improve modelling of global-context without relying on strong spatial priors. Subsequent modifications to this vanilla Transformer \cite{A.Vaswani2017} have improved the architecture for vision applications: by modelling contextual information at different scales
\cite{B.Heo2021,CF.Chen2021}, revisiting self-attention with regional or local attention \cite{Ze.Liu2021,C.Chen2022}, or re-introducing some inductive biases \cite{S.Dascoli2021}. Such methods have returned state-of-the-art performances for many image or video understanding tasks \cite{H.Touvron2020, Ze.Liu2021,X.Zhu2020,K.Li2022,C.Chen2022}, where some of this success may be attributed to the scalability of ViT to be pre-trained on very large datasets \cite{ViT}, but also to efficient (pre-)training schemes \cite{H.Touvron2020,J.Zihang2021,M.Caron2021,Bao2021}.

Similarly, transformer model \cite{A.Vaswani2017} and self-attention modules have been adapted to improve the context-modelling over gDL methods for non-Euclidean manifolds: in point-clouds \cite{Engel2020,Qin2022}, shape \cite{Yan2022}, or graphs networks \cite{Yun2019,Kim2021}. 

In this paper, we propose the Surface Vision Transformer (\emph{SiT}), a methodology for modelling functions on surfaces by extending the ViT to surface meshes through proposing a mechanism for surface patching. To do so, surface data are projected onto a sphere and patched using a regular icospheric tessellation. This reformulates any surface learning task that can adapt to genus-zero surfaces as a sequence-to-sequence problem. We validate this methodology for two cortical phenotype regression tasks, against number of geometric deep learning (gDL) methods, benchmarked in \cite{A.Fawaz2021} on cortical phenotype regression and segmentation. The key contributions of this paper are as follows:

\begin{itemize}
    \item  We introduce a framework for sequence-to-sequence modelling of surfaces, which patch surfaces via projection to a regularly tessellated icosphere. 
    \item Surface Vision Transformers (\emph{SiT}) are compared against geometric CNNs, benchmarked in \cite{A.Fawaz2021}, and demonstrate superior performance for regression of developmental phenotypes.
    \item \emph{SiT} also exhibits some degree of transformation invariance by performing closely on registered and unregistered scans, without incorporating strong inductive bias in the architecture. 
\end{itemize}

\section{Methods}

\begin{figure}[h]
  \centering
\makebox[\linewidth]{
	\includegraphics[width=1.0\columnwidth]{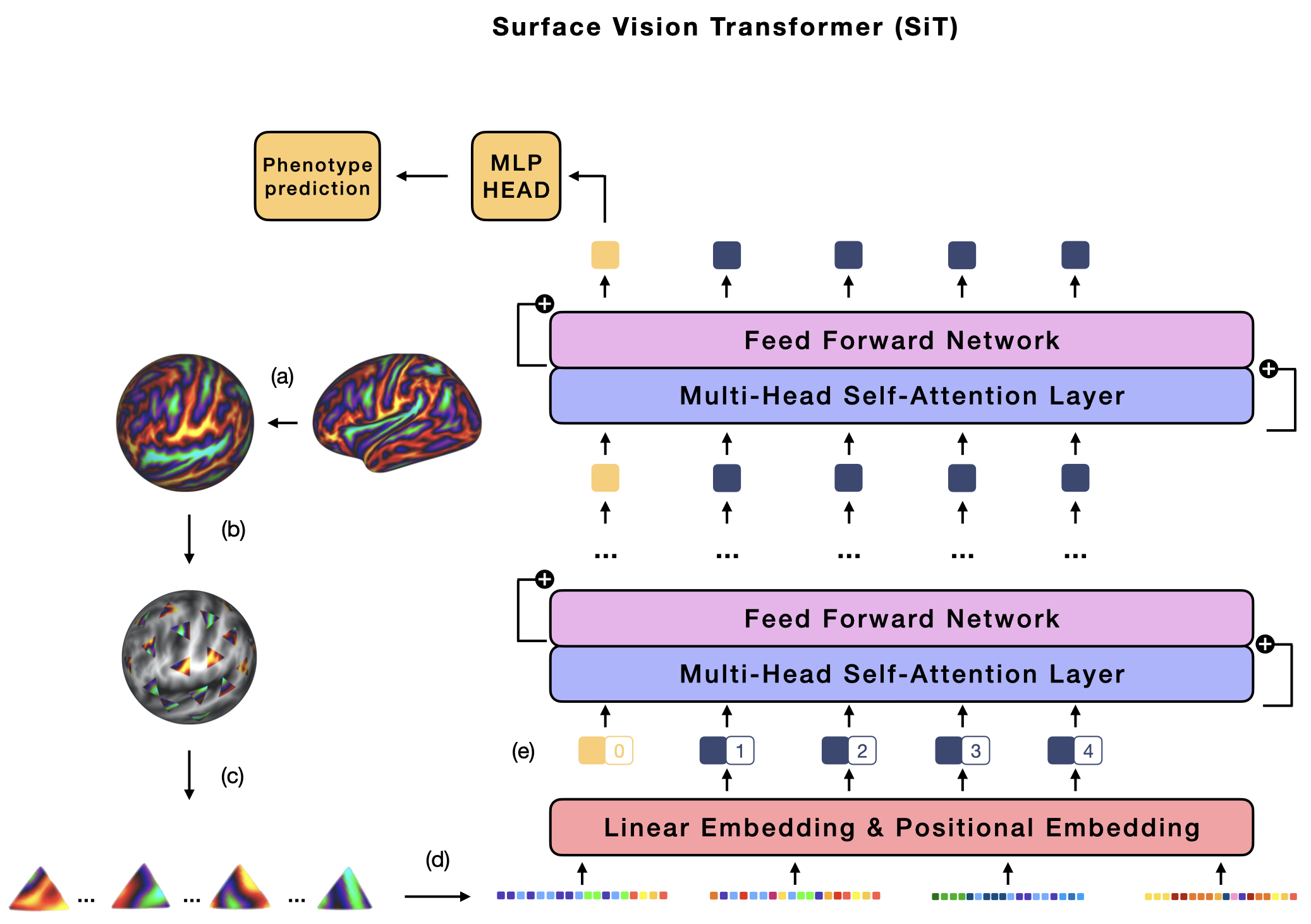}}
\caption{The cortical data is first resampled (a), using barycentric interpolation, from its template resolution (32492 vertices) to a sixth order icosphere (mesh of 40962 equally spaced vertices). The regular icosphere is divided into triangular patches of equal vertex count (b, c) that fully cover the sphere (not shown), which are flattened into feature vectors (d), and then fed into the transformer model.}
\label{figure:sit}
\end{figure}

\subsection{Architecture} 

The \emph{SiT} model translates surface understanding to a sequence-to-sequence learning task by reshaping the high-resolution grid of the input domain $X$,  into a sequence of $N$ flattened patches $ \widetilde{X} = \left [ \widetilde{X}^{(0)}_1, ..., \widetilde{X}^{(0)}_N \right] \in \mathbb{R}^{N\times (VC)}$ (V vertices, C channels). These are first projected onto a $D-$dimensional sequence $X^{(0)} = \left [ X^{(0)}_1, ..., X^{(0)}_N \right]\in \mathbb{R}^{N\times D}$, using a trainable linear layer. Then, an extra $D$-dimensional token for regression is concatenated ($X^{(0)}_0$), and a positional embedding ($E_{pos} \in \mathbb{R}^{(N+1)\times D}$) is added, such that the input sequence of the transformer becomes $ X^{(0)} = \left [ X^{(0)}_0, ..., X^{(0)}_N \right] + E_{pos}$ (see Fig \ref{figure:sit}(b-e)). The sequence of embeddings is then processed by a vanilla Transformer encoder as in \cite{ViT} with consecutive transformer encoder blocks of \textit{Multi-Head Self-Attention} (MHSA) and \textit{Feed Forward Network} (FFN) layers, with residual layers in-between. The architecture of the \emph{SiT} is illustrated in Figure \ref{figure:sit}.
Here, the proposed \emph{SiT} model builds upon two variants of the data efficient image transformer or \textit{DeiT} \cite{H.Touvron2020}: \emph{DeiT-Tiny}, \emph{DeiT-Small}, adapted into smaller versions from the vanilla Vision Transformer (ViT) \cite{ViT}.
A number of $L=12$ layers or transformer encoder blocks is used for both SiT versions; however, they differ in their number of heads, hidden size or embedding dimension $D$, and in the number of neurons (\emph{MLP size}) in the FFN, details in Table \ref{tab:vit-size}.


\subsection{Surface Patching}
\label{sec-patching}

The \emph{SiT} can generate patches from any regularly tessellated reference grid that supports down-sampling. For the cortical surface, this is achieved by imposing a low-resolution triangulated grid on the input mesh, using a regularly tessellated icosphere (Fig \ref{figure:sit}(b)).

Here, cortical surface data were first projected to a regularly-tessellated sphere (with 32,492 vertices) as part of the dHCP structural pipeline \cite{A.Makropoulos2018}. Spherical data were then resampled onto a regularly-tessellated sixth-order icosahedron with 40,962 vertices, then split into triangular patches where each patch corresponds to all data points within one face of a second-order icosphere (153 vertices per patch). The sequence is thus made of 320 non-overlapping patches that only share common edges (Fig.\ref{figure:sit} (a-c)).


\subsection{Optimisation}

To mitigate the lack of inductive biases in the architecture, transformers typically require large training datasets or efficient (pre-)training strategies \cite{ViT, Bao2021, H.Touvron2020, A.Steiner2021}. Therefore, we explore techniques for improving model generalisation in a context of a neuroimaging dataset of limited size: specifically pre-training and augmentation.

\paragraph{Pre-training} is relevant for biomedical imaging tasks, as datasets are usually smaller than in natural imaging, and can benefit from pre-training before transferring to downstream tasks. In this paper we evaluate different training strategies: 1) training from scratch; 2) initialising from ImageNet weights (to support training on small datasets through incorporation of some  spatial priors) and 3) fine-tuning after BERT-like pretraining, a well-known self-supervised pretraining strategy. For ImageNet, we used pretrained models from the \textbf{timm} open-source library\footnote{pretrained models on ImageNet available at \url{http://github.com/rwightman/pytorch-image-models/}}, where models were pretrained on ImageNet2012 (1 million images, 1000 classes) on patches of size $16 \times 16 \times 3$. Self-supervision is implemented as a \emph{masked patch prediction} (MPP) task, following the approach proposed in BERT \cite{J.Devlin2019}, which consists of corrupting at random some input patches in the sequence; then training the network to learn how to reconstruct the full corrupted patches. In this setting, we corrupt at random 50\% of the input patches, either replacing them with a learnable mask token (80\%), another patch embedding from the sequence at random (10\%) or keeping their original embeddings (10\%). To optimise the reconstruction, the mean square error (MSE) loss is computed only for the patches in the sequence that were masked. 

\paragraph{Data augmentation} Following previous work \cite{S.Dahan2022}, we additionally propose to augment the icosahedral patch selection by implementing $\pm \left \{5^{\circ},10^{\circ},15^{\circ},20^{\circ},25^{\circ},30^{\circ} \right \}$ rotations of the sphere before patching around one of the x,y,z axes. Dropout was also used before the transformer encoder and inside the FFN networks, compared to \cite{S.Dahan2022}.

\begin{table}[H]
  \centering
  \small
  \setlength{\tabcolsep}{2pt}
  \begin{tabular}{lccccc}
    \toprule
    \textbf{Models} & Layers & Heads & Hidden size \textit{D} & MLP size  & Params.\\
    \midrule
    SiT-Tiny & 12 & 3 & 192 & 768 & 5.5M \\
    SiT-Small & 12 & 6 & 384 & 1536 & 21.6M \\
    \bottomrule
  \end{tabular}
  \caption{Architectures of \emph{SiT-tiny} and \emph{SiT-small}, inspired by DeiT models in \cite{H.Touvron2020}}
  \label{tab:vit-size}
\end{table}

\section{Experiments \& Results}

\begin{table*}[h]
  \footnotesize
  \centering
  \setlength{\tabcolsep}{7pt}
  \renewcommand{\arraystretch}{1.1}
  \begin{tabular}{l|c|c|c|c|c|c|c|c}
    \hline
    \multirow{2}{1.2cm}{\textbf{Methods}} &\multirow{2}{1.5cm}{\textbf{Pretraining}}  &\multicolumn{3}{c|}{\textbf{PMA}} & \multicolumn{3}{c|}{\textbf{GA - deconfounded}}  &\multirow{2}{1.0cm}{\textbf{Average}}\\
    \cline{3-8}
    &&  \textbf{Template} & \textbf{Native}&   \textbf{Avg}&  \textbf{Template} &\textbf{Native} & \textbf{Avg} \\
    \hline
    \hline
    S2CNN &\xmark & 0.63 $\pm0.02$& 0.73 $\pm0.25$ & 0.68& 1.35 $\pm0.68$& 1.52 $\pm0.60$&1.44&1.06\\
    ChebNet &\xmark   & 0.59 $\pm0.37$ & 0.77 $\pm0.49$&0.68& 1.57 $\pm0.15$& 1.70 $\pm0.36$&1.64& 1.16\\
    GConvNet &\xmark    & 0.75 $\pm0.13$& 0.75 $\pm0.26$&0.75& 1.77 $\pm0.26$& 2.30 $\pm0.74$&2.04&1.39\\
    Spherical UNet & \xmark   & 0.57 $\pm0.18$& 0.87 $\pm0.50$&0.75& \textbf{0.85} $\pm0.17$& 2.16 $\pm0.57$&1.51&1.11\\
    MoNet & \xmark  & 0.57 $\pm0.02$ & \textbf{0.61} $\pm0.05$& \textbf{0.59}& 1.44 $\pm0.08$&1.58 $\pm0.06$&1.51&1.05\\
    \hline
    \hline
    SiT-tiny &\xmark &0.63 $\pm0.01$ &0.77 $\pm$0.03&0.70&1.17 $\pm$0.04 & 1.36 $\pm$0.01&1.27&0.98\\
    SiT-tiny &\textit{ImageNet}& 0.67 $\pm$0.02& 0.70 $\pm$0.04 &0.69&1.11 $\pm$0.02&1.20 $\pm$0.10&1.16&0.92\\
    SiT-tiny &\textit{MPP}&  0.58 $\pm$0.01 & 0.64 $\pm$0.06&0.61& 1.03 $\pm$0.09 & 1.31 $\pm$0.01&1.17&0.89\\
    \hline
    SiT-small &\xmark&  0.60 $\pm$0.02& 0.75 $\pm$0.01  &0.68&1.14 $\pm$0.05& 1.22 $\pm$0.04&1.18&0.93\\
    SiT-small &\textit{ImageNet}&  0.59 $\pm$0.03 & 0.71 $\pm$0.02& 0.65&1.13 $\pm$0.03&1.30 $\pm$0.08&1.22&0.93\\
    SiT-small &\textit{MPP}& \textbf{0.55} $\pm$0.04& 0.63 $\pm$0.06&\textbf{0.59}&1.02 $\pm$0.06& 1.21 $\pm$0.12&\textbf{1.12}&\textbf{0.85}\\
    \hline
    \hline
  \end{tabular}
  \caption{Results of \emph{SiT-tiny} and \emph{SiT-small} models for the task of PMA and GA on template and native space, for three training configurations. Best MAE on the test set and standard deviations (over three trainings) are reported. Two configurations of the data were used: \emph{template} where data are aligned and \emph{native} (unregistered).}  
  \label{tab:results-vit}
\end{table*}

We evaluate the performance of \emph{SiT} on two challenging tasks using neonatal cortical surface data from the developing Human Connectome Project (dHCP) \cite{E.Hughes2017}; 1) prediction of postmenstrual age at scan (PMA), and 2) gestational age at birth (GA).
 Stochastic gradient descent (SGD) with momentum was used for model optimisation, compared to Adam optimisation for gDL models \cite{A.Fawaz2021}. \emph{SiT} models were trained for 2000 iterations from scratch and only 1000 iterations following pre-training as convergence appeared to be faster. All experiments were run on a single NVIDIA RTX3090 24GB GPU. A batch size of 256 was used for \emph{SiT-tiny} and 128 for \emph{SiT-small}.

\paragraph{Data \& Training}

Data for this experiment corresponds to cortical surface data from the third release of the developing Human Connectome Project (dHCP) \cite{E.Hughes2017}. Surfaces were extracted from T2- and T1-weighted Magnetic Resonance Imaging (MRI) scans using the dHCP structural pipeline \cite{A.Makropoulos2018, MKuklisova-Murgasova2012, A.Schuh2017,E.Hughes2017, L.Cordero-Grande2018}. Four cortical surface metrics were used: sulcal depth, curvature, cortical thickness and T1w/T2w ratio (intracortical myelination). Data were registered using Multimodal Surface Matching \cite{E.Robinson2014,E.Robinson2018} to the left-right symmetric dHCP spatiotemporal cortical atlas \cite{J.Bozek2018,L.Williams2021}.

A total of 588 images were included, acquired from term (born $\ge37$ weeks gestational age, GA) and preterm (born $<37$ weeks GA) neonatal subjects, scanned between 24 and 45 weeks postmenstrual age (PMA). Some of the preterm neonates were scanned twice: once after birth and again around term-equivalent age. The proposed framework was benchmarked on two phenotype regression tasks:  prediction of postmenstrual age (PMA) at scan, and gestational age (GA) at birth, where since the objective was to model PMA and GA as markers of healthy development, all preterms' second scans were excluded from the PMA prediction task, and all first scans were excluded from the GA prediction task. This resulted in 530 neonatal subjects for the PMA prediction task (419 term/111 preterm), and 514 subjects (419 term/95 preterm) for the GA prediction task. 
The dHCP dataset is heavily unbalanced with more term babies than preterm babies. In extension to previous work \cite{S.Dahan2022}, this class imbalance was addressed by adapting sampling during training. Subjects were split into 3 categories, which reflect the clinical subcategories of preterm birth \cite{spong2013defining}: over 37 weeks, between 32 and 37 and below 32 weeks. The original ratio of examples in each of these three categories was 1/7/11. Experiments were run on both \emph{template}-aligned data and unregistered (\emph{native}) data, and train/test/validation splits parallel those used in \cite{A.Fawaz2021}.  

Changes to cortical organisation are implicated in numerous neurological and developmental disorders \cite{raznahan2010cortical,hong2018multidimensional}. Such disorders are diffuse processes, heterogeneous between individuals and populations and cannot be studied effectively using traditional approaches (based on spatial normalisation to global average template) since human brains vary in ways that violate the assumptions of traditional image registration \cite{M.Glasser2016}, and thereby limit the sensitivity of population-based comparisons. This motivates the use of \emph{SiT} as an attention-modelling tool for cortical analysis on two neonatal imaging tasks that exhibit high variability in cortical development between subjects. 

\paragraph{Deconfounding strategy} The task of GA prediction is arguably more complicated than the PMA task, as it is run on scans acquired around term-equivalent age (37-45 weeks PMA) for both term and preterm neonates, and therefore is highly correlated to PMA at scan. Here, a deconfounding strategy was employed, following \cite{S.Dahan2022}, where the scan age information was incorporated into the patch sequence by adding an extra embedding to all patches in the sequence before the transformer encoder. This was implemented using a fully connected network to project scan age to a vector embedding of dimension D after batch-normalisation.

\paragraph{Results}
The proposed \emph{SiT} models were compared against the best performing surface CNNs reported in \cite{A.Fawaz2021}: Spherical U-Net \cite{F.Zhao2019}, MoNet \cite{F.Monti2017}, GConvNet \cite{T.Kipf2017}, ChebNet \cite{M.Defferrard2017} and S2CNN \cite{T.Cohen2018} (Table \ref{tab:results-vit}). We should stress that these gDL models were trained with both rotational and non-linear data augmentations \cite{A.Fawaz2021}. 

Overall, \emph{SiT-small} and \emph{SiT-tiny} configurations consistently outperformed three of the gDL methods (S2CNN, GConvNet, and ChebNet) for all tasks. On average, the 6 \emph{SiT} configurations achieved prediction errors below 0.98 MAE, compared to 1.05 MAE for the best gDL model on average: MoNet. Best performance overall (0.85 MAE across tasks) was obtained with \emph{SiT-small} pre-trained with MPP, with  large improvement for GA prediction: 1.12 MAE (on average template \& native) against 1.44 MAE for S2CNN.

For the task of PMA, \emph{SiT-small} pretrained obtained performances on template and native data (0.55/0.63) comparable to the best gDL model MoNet: 0.57/0.61. The use of dropout and rotation augmentation did not seem to improve the performances of \emph{SiT} for the task of PMA, which already achieved good results without regularisation, whereas augmentation (specifically rotations $\pm \left \{5^{\circ},10^{\circ} \right \}$) and dropout greatly improved \emph{SiTs'} performance for GA prediction where \emph{SiT} models outperformed gDL methods for all native configuration and template configuration (except for Spherical UNet-template that under-performs greatly in native space). 

Across all tasks, \emph{SiTs} demonstrate consistent performance across training runs with smaller variability compared to surface CNNs. The methodology also demonstrates robustness between template-aligned and native data, dropping less in performance than some gDL methods, such as Spherical UNet which obtained 0.85 MAE on GA-template but does not build rotational equivariance (2.16 MAE on GA-native). All \emph{SiTs} also outperform MoNet (the best gDL method) for both GA-template and GA-native, which although rotationally equivariant and consistent between native and template, learns less expressive convolutional filters (parameterised as a mixture of Gaussians).
Finally, pre-training generally improves performances of \emph{SiTs} compared to training from scratch. This is the case for all \emph{SiTs} trained following the MPP self-supervision task, and for 6/8 configurations of \emph{SiTs} following ImageNet initialisation, but with slighter improvements in the later case.

\section{Discussion}

In this paper, we demonstrated that surface understanding is possible with vision transformers. This was obtained by introducing a patching methodology for surface data that can be projected onto a spherical manifold. The \emph{SiT} methodology surpasses in performance many geometric deep learning methods in the context of cortical analysis, showing some degree of transformation invariance with far less drop in performance on unregistered data than most performing gDL frameworks, and greatly improved by training with augmentations, comparatively to \cite{S.Dahan2022}.


The use of vision transformers constitutes an exciting opportunity for many surface learning applications, especially in the context of biomedical data to study diffuse processes in cardiac \cite{C.Mauger2019}, or neurodevelopmental modelling \cite{R.Dimitrova2021}; and where surface deep learning models are usually limited by the receptive field of convolution operations.
Various improvement of the method could be explored as the \emph{SiT} only employs a vanilla Transformer encoder \cite{A.Vaswani2017}. Latest developments around multi-scale feature learning in ViT \cite{B.Heo2021,CF.Chen2021,Ze.Liu2021,C.Chen2022} would further benefit the context-modelling of cortical surface, as new (pre-)training schemes \cite{H.Touvron2020,A.Steiner2021}.




{\small
\bibliographystyle{ieee_fullname}
\bibliography{cvprbib}
}

\end{document}